\documentclass[12pt,preprint]{revtex4}
\usepackage{graphicx, epsfig, rotate}

\begin{document}

\title{Qualitative study of a robot arm as a hamiltonian system}

\date{\today}

\author{G. A. Monerat}
\email{monerat@if.uff.br}
\affiliation{Universidade Federal Fluminense, Instituto de
F\'{\i}sica, Av. Litor\^{a}nea s/n$^o$, Boa Viagem, 
Niter\'{o}i --- RJ, 24210-340, Brasil}

\author{E. V. Corr\^{e}a Silva}
\email{eduvasquez@bol.com.br}
\affiliation{Universidade Federal Fluminense, Instituto de
F\'{\i}sica, Av. Litor\^{a}nea s/n$^o$, Boa Viagem, 
Niter\'{o}i --- RJ, 24210-340, Brasil}

\author{A. G. Cyrino} 
\email{aline.gal@uol.com.br} 
\affiliation{\small
Universidade do Estado do Rio de Janeiro, Faculdade de Engenharia,
Campus Regional de Resende. Estrada Resende Riachuelo, s/n$^o$. Morada
da Colina, Resende --- RJ,  27523-000, Brasil}


\begin{abstract}

A double pendulum subject to external torques is used as a model to
study the stability of a planar manipulator with two links and two
rotational driven joints. The hamiltonian equations of motion and the
fixed points (stationary solutions) in phase space are determined.
Under suitable conditions, the presence of constant torques does not
change the number of fixed points, and preserves the topology of
orbits in their linear neighborhoods; two equivalent invariant
manifolds are observed, each corresponding to a saddle-center fixed
point.

\end{abstract}

\maketitle

\section{Introduction}

\begin{figure}[htb]
\includegraphics[width=3.9in,height=3.5in]{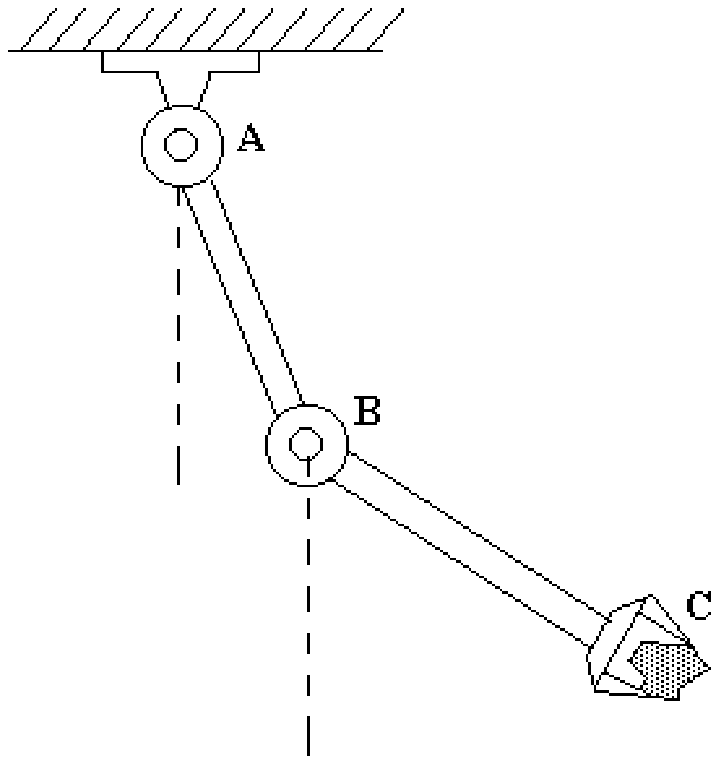}
\caption{\label{f1}\small Simplified diagram of a planar manipulator with
	two links.}
\end{figure}

\begin{figure}[htb]
 \includegraphics[width=10.0 cm,height=8.9 cm]{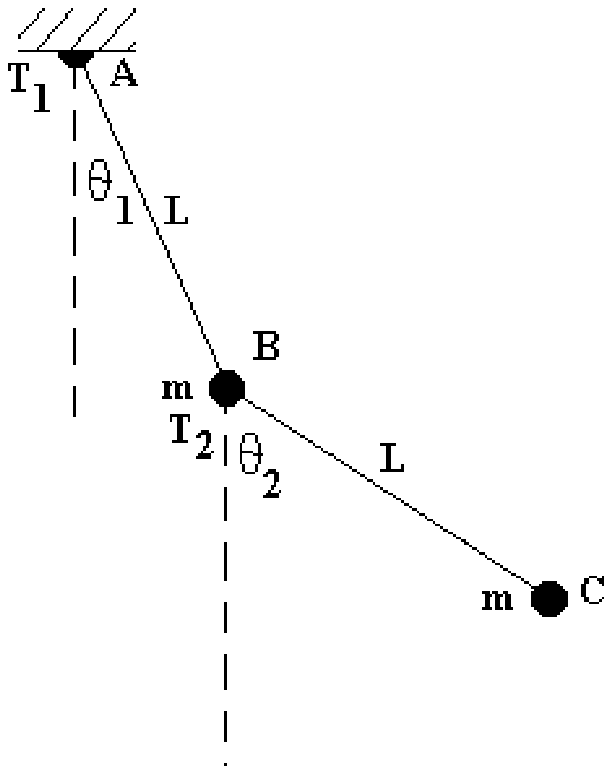}
\caption{\label{f2}\small Double pendulum model for the planar manipulator
	in Fig.\ref{f1}. The pendulum is subject to the external torques $T1$
	and $T2$ at points $A$ and $B$, respectively.}
\end{figure}

\indent The problem of the stability of motion and equilibrium of
manipulators is essential to their applicability in industry. In this
work, we analyze the stability of one class of such manipulators,
namely, planar manipulators with two links and two rotational driven
joints (see Fig.\ref{f1}), modelled by a double pendulum subject to
two constant external torques $T_1$ and $T_2$(see Fig.\ref{f2}). Each
pendulum has lenght $L$ and has a particle of mass $m$ attached to its
end.

For any given configuration $(\theta_1, \theta_2)$ where $0 \leq
\theta_1 \leq 2\pi$ and $0 \leq \theta_1 \leq 2\pi$, and conjugate
momenta $p_1=p_2=0$, the torques $T_1$ and $T_2$ can be adjusted so
that this configuration become stationary. In this work we will
study the dynamics of this system in the linear neighborhod of
stationary configurations under constant external torques.

As indicated in Fig.\ref{f2}, the model consists of a combination of
two simple pendula with equal length $L$, where the two drivers are
represented by external torques $T_1$ and $T_2$ applied at points $A$
and $B$, respectively. For the sake of simplicity, we shall consider
the mass $m$ of the object held by the end-effector $C$ as being equal to
the mass of the driver in $B$. 

We apply qualitative analysis
techniques\cite{ahmed,hagedorn,meirovitch} suitable to non-linear
systems, using the hamiltonian formalism.

\section{Dynamics of the model}

\indent The system depicted in Fig.\ref{f2} can be represented by the
following hamiltonian function with two degrees of freedom:
\begin{eqnarray}
H &=& 
	\frac{1}{mL^2\left(1+sen^2\left({\theta}_{1}-{\theta}_{2}\right)\right)}
	\left\{\frac{p_{1}^2}{2}+p_{2}^2-cos\left({\theta}_{1}-{\theta}_
	{2}\right)p_{1}p_{2}\right\} + \nonumber \\
&  & - mgL\left(\cos({\theta}_{1})+2 \cos({\theta}_{2})\right)
	-{\theta}_{1}T_{1}(t)-{\theta}_{2}T_{2}(t),
	\label{hamiltoniana}
\end{eqnarray}

\noindent where $g$ is the acceleration of gravity and $(p_{1},p_{2})$
are the conjugate momenta to the angular variables
$({\theta}_{1},{\theta}_{2})$, respectively. The functions $T_1(t)$
and $T_2(t)$ stand for arbitrary external torques on the system. The
hamiltonian equations of motion yield the non-integrable, non-linear
dynamical system

\begin{eqnarray}
\dot{{\theta}_{1}} = \frac{\partial H}{\partial {p}_{1}} &=&
	\frac{p_{1}-\cos({\theta}_{1}-{\theta}_{2})p_{2}}{mL^2
	\left(1+\sin^2({\theta}_{1}-{\theta}_{2})\right)}
	\label{eqshamilton1} \\ 
\dot{p_{1}} = -\frac{\partial H}{\partial {\theta}_{1}} &=&
	\frac{\left(p_{1}^2+2p_{2}^2-2\cos({\theta}_{1}-{\theta}_{2})p_{1}p_{2}
	\right)\cos(\theta_1-\theta_2)\sin({\theta}_{1}-{\theta}_{2})}{mL^2\left
	(1+\sin^2({\theta}_{1}-{\theta}_{2})\right)^2} + 
	\nonumber \\
& & - \frac{\sin({\theta}_{1}-{\theta}_{2})p_{1}p_{2}}
	{mL^2(1+\sin^2({\theta}_{1}-{\theta}_{2}))}
	-2mgL\sin({\theta}_{1})+T_{1}(t)
	\label{eqshamilton2} \\
\dot{{\theta}_{2}} = \frac{\partial H}{\partial {p}_{2}} &=&
	\frac{2p_{2}-\cos({\theta}_{1}-{\theta}_{2})p_{1}}{mL^2
	\left(1+\sin^2({\theta}_{1}-{\theta}_{2})\right)}
	\label{eqshamilton3} \\
\dot{p_{2}} = -\frac{\partial H}{\partial {\theta}_{2}}&=&
	-\frac{\left(p_{1}^2+2p_{2}^2-2\cos({\theta}_{1}-{\theta}_{2})p_{1}p_{2
	}\right)\cos({\theta}_{1}-{\theta}_{2})\sin({\theta}_{1}-{\theta}_{2})}{
	mL^2\left(1+\sin^2({\theta}_{1}-{\theta}_{2})\right)^2}+
	  \nonumber \\
	& & + \frac{\sin({\theta}_{1} - {\theta}_{2}) p_{1} p_{2}}{mL^2\left(1 +
	\sin^2({\theta}_{1} - {\theta}_{2})\right)}-
	mgL\sin({\theta}_{2})+T_{2}(t)
	\label{eqshamilton4}.
\end{eqnarray} 

\vspace{0.5cm}

We will search for fixed points in phase space. At first, we will let
$T_1 = T_2 = 0$ and discuss the case of null external torques;
afterwards, we shall treat the case of non-null constant external
torques.

\section{The robot arm with null external torques}

Hamilton equations (\ref{eqshamilton1}) and (\ref{eqshamilton4}) have
four fixed points. Namely,
\begin{equation}
{\cal P}_{1} = 
	\left(
	\begin{tabular}{c}
	0 \\
	0 \\
	0 \\
	0
	\end{tabular}
	\right) ,\ \
{\cal P}_{2} = 
	\left(
	\begin{tabular}{c}
	0 \\
	0 \\
	$\pi$ \\
	0
	\end{tabular}\right) ,\ \ 
{\cal P}_{3} = 
	\left(
	\begin{tabular}{c}
	$\pi$ \\
	0 \\
	0 \\
	0
	\end{tabular}\right) ,\ \ 
{\cal P}_{4} = 
	\left(
	\begin{tabular}{c}
	$\pi$ \\
	0 \\
	$\pi$ \\
	0
	\end{tabular}
	\right).
\end{equation}

\noindent where the matrix lines are ordered from top to bottom
according to the coordinates $(\theta_1, p_1, \theta_2, p_2)$,
respectively. The total energies associated to each fixed point are
\begin{equation}
E_{1} = -3mgL \mbox{\hspace{5mm} , \hspace{5mm}}
E_{2} = -mgL \mbox{\hspace{5mm} , \hspace{5mm}}
E_{3} = +mgL \mbox{\hspace{5mm} , \hspace{5mm}}
E_{4} = +3mgL.
\end{equation}

\noindent Upon linearization of the system of Hamilton equations, we obtain
\begin{equation}
\dot{X}_{i} = J_{i}X_i, 
\label{sistemalinearizado}
\end{equation}

\noindent where $i=1,2,3,4$ labels the fixed points. 
The vector $X_{i}$ has the general form 
$\tilde{X}_i = (\theta_1, p_1, \theta_2, p_2)$, and the jacobian matrices
$J_{i}$
are 
\begin{equation}
J_{1} =\left(
\begin{array}{cccc}
0 & \frac{1}{mL^2} & 0 & - \frac{1}{mL^2} \\
-2mgL & 0 & 0 & 0 \\
0 & -\frac{1}{mL^2} & 0 & \frac{2}{mL^2} \\
0 & 0 & -mgL & 0 \\
\end{array}
\right),
\label{jacobiana1}
\end{equation}

\begin{equation}
J_{2} =\left(
\begin{array}{cccc}
0 & \frac{1}{mL^2} & 0 & \frac{1}{mL^2} \\
-2mgL & 0 & 0 & 0 \\
0 & \frac{1}{mL^2} & 0 & \frac{2}{mL^2} \\
0 & 0 & mgL & 0 \\
\end{array}
\right),
\label{jacobiana2}
\end{equation}

\begin{equation}
J_{3} =\left(
\begin{array}{cccc}
0 & \frac{1}{mL^2} & 0 &  \frac{1}{mL^2} \\
2mgL & 0 & 0 & 0 \\
0 & \frac{1}{mL^2} & 0 & \frac{2}{mL^2} \\
0 & 0 & -mgL & 0 \\
\end{array}
\right),
\label{jacobiana3}
\end{equation}

\begin{equation}
J_{4} =\left(
\begin{array}{cccc}
0 & \frac{1}{mL^2} & 0 & -\frac{1}{mL^2} \\
2mgL & 0 & 0 & 0 \\
0 & -\frac{1}{mL^2} & 0 & \frac{2}{mL^2} \\
0 & 0 & mgL & 0 \\
\end{array}
\right).
\label{jacobiana4}
\end{equation}

\noindent The general solution to the linearized system
(\ref{sistemalinearizado}) around a fixed point is a superposition of
four independent solutions,
\begin{equation}
X_i(t) = \sum_{m=1}^{4} c^{(i)}_{m} e^{\lambda^{(i)}_{n} t} A^{(i)}_{m},
\label{solucaolinear}
\end{equation}

\noindent where $A^{(i)}_{m}$ are the eigenvectors associated to the
eigenvalues ${\lambda}^{(i)}_{n}$ of $J_i$, and the coefficients
$c^{(i)}_n$ are integration constants that depend on the initial
conditions. The eigenvalues associated to the $J_{i}$ matrices are,
respectively,
\begin{eqnarray}
&J_{1}: & 
	\lambda^{(1)}_{1,2}=\pm i \omega_{0} \sqrt{2-\sqrt{2}} 
	\mbox{\hspace{5mm} and \hspace{5mm}} 
	{\lambda}^{(1)}_{3,4}=\pm i \omega_{0} \sqrt{2+\sqrt{2}} ~;
	\label{autovalores1} \\ 
& & \nonumber \\
&J_{2} :& 
	{\lambda}^{(2)}_{1,2}=\pm  \omega_{0} \sqrt[4]{2}
	\mbox{\hspace{5mm} and \hspace{5mm}} 
	{\lambda}^{(2)}_{3,4}=\pm i \ \omega_{0} \sqrt[4]{2} ~;
	\label{autovalores2} \\ 
& & \nonumber \\
&J_{3} :&
	{\lambda}^{(3)}_{1,2}=\pm \omega_{0} \sqrt[4]{2}
	\mbox{\hspace{5mm} and \hspace{5mm}} 
	{\lambda}^{(3)}_{3,4}=\pm i \ \omega_{0} \sqrt[4]{2} ~;
	\label{autovalores3} \\
& & \nonumber \\
&J_{4} :& 
	{\lambda}^{(4)}_{1,2}=\pm \omega_{0} \sqrt{2-\sqrt{2}} 
	\mbox{\hspace{5mm} and \hspace{5mm}} 
	{\lambda}^{(4)}_{3,4}=\pm \omega_{0} \sqrt{2+\sqrt{2}}~.
	\label{autovalores4}
\end{eqnarray}

\noindent where $\omega_0=\sqrt{g/L}$ is the natural oscillation
frequency of as simple pendulum for small amplitudes. In view of
eqs.(\ref{autovalores1}) to (\ref{autovalores4}), we may
classify\cite{stuchi,monerat} the four existing fixed points: ${\cal
P}_{1}$ is a pure center; ${\cal P}_{2}$ and ${\cal P}_{3}$ are
saddle-centers and ${\cal P}_{4}$ is a pure saddle.

\subsection{The invariant manifolds: case of null torques}

\indent An important characteristic of this system when the torques
are null is the presence of two similar invariant manifolds, each one
of them associated to a saddle-center fixed point. Those manifolds are
\begin{eqnarray}
{\cal M}_{1} & : & \left(\theta_2=0, 
	p_2= \frac{p_1}{2}\cos \theta_1 \right)
	\label{manifold1}\\ 
& & 	\nonumber  \\
{\cal M}_{2} & : & \left(\theta_2=\pi,
	p_2= -\frac{p_{1}}{2}\cos \theta_1 \right).
	\label{manifold2}
\end{eqnarray}


\vspace{-0.5cm}

\begin{figure}[htb]
\includegraphics{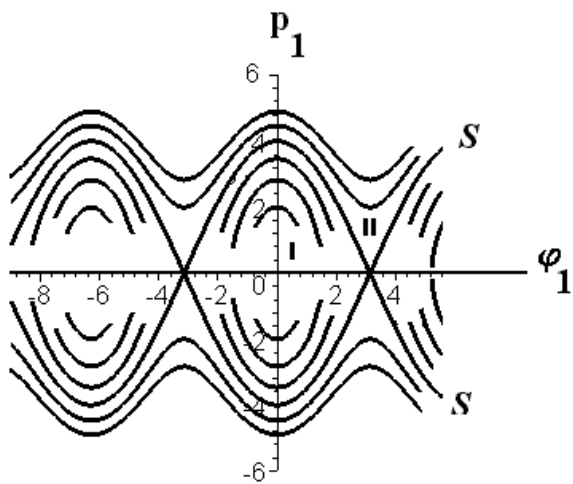}
\caption{\label{retrato1}\small Phase portrait on the invariant
	manifold ${\cal M}_{1}$.}
\end{figure}

In view of the resemblance of the two invariant manifolds, our
discussion will focus on ${\cal M}_{1}$, but can be straighforwardly
extended to ${\cal M}_{2}$. The phase portrait upon ${\cal M}_{1}$ is
shown in Fig.\ref{retrato1}.  The
resulting phase portrait is equivalent to that of a mathematical
pendulum with arbitrary amplitude of oscillation.

\noindent Dynamics upon ${\cal M}_{1}$ is governed by the
two-dimensional system of equations

\begin{equation}
\left\{
\begin{array}{lll}
\dot{{\theta}}_{1} =& \frac{\displaystyle \partial H}{\displaystyle
	\partial p_{1}}= \frac{\displaystyle p_{1}}{\displaystyle 2mL^2}\\ 
& \\
\dot{p}_{1} =& -\frac{\displaystyle \partial H}{\displaystyle \partial
	q_{1}}= -2mgL\sin({\theta}_{1}).\\ 
\end{array}
\right.
\end{equation}

\noindent In Fig.\ref{retrato1} each orbit has a definite total
energy. The point I corresponds to the solution $(\theta_1=0, p_1=0)$,
a pure center upon the invariant manifold. That indicates that the
robot arms is standing still in the vertical position. Another
saddle-center fixed point upon ${\cal M}_{1}$ is
$(\theta_1=\pi,p_1=0)$. They correspond to the case where the $AB$
link of the robot arm (cf.Figs.\ref{f1} and \ref{f2}) remains in the
vertical position (downwards and upwards, respectively), and only the
$BC$ link is free to move.

\subsection{The normal form of the Hamiltonian}

\indent The hamiltonian describing the dynamics in the linear
neighborhood of a saddle-center can always be rewritten as the sum of
a {\it rotational energy} term and a {\it hyperbolic energy} term. This is a
consequence of Moser's theorem \cite{moser}, which establishes that in
a sufficiently small neighborhood of any saddle-center point, there
is a set of canonically conjugate variables so that the
hamiltonian in that neighboorhood is separable into a purely
rotational term, and a purely hyperbolic term. 
We will apply the method of normal forms to find this set of
coordinates.

\indent The method of normal forms\cite{osorio} consists basically in
applying Taylor series expansion to the velocity field around a fixed
point. When such point is a saddle-center point, we may write the
expanded hamiltonian in its quadratic (or {\it normal}) form. A
canonical transformation is carried out so that we have vanishing
off-diagonal terms\cite{landau}. The transformation for the linear
neighborhood of ${\cal P}_{2}$ is
\begin{eqnarray}
\theta_1 & = & 
	x+y ~; 
	\label{canonica1}\\
p_1 & = & 
	-\frac{ 2\left(5+\sqrt{17}\right)}{ 17+5\sqrt{17}}\left(\frac{ 2p_{
	x}}{ 5+\sqrt{17}}-\frac{\left( 5+\sqrt{17}\right) p_{ y}}{
	4}\right)~;
	\label{canonica2}
	\\
\theta_2= &  & 
	\frac{\left( 5+\sqrt{17}\right)}{ 4}x+\frac{ 2}{ 5+\sqrt{17}}y+\pi~;
	\label{canonica3}
	\\
p_2 & = &
	\frac{ 2\left(5+\sqrt{17}\right)}{17+5\sqrt{17}}\left(p_{x}-p_{y}\right).
\label{canonica4}
\end{eqnarray}

\noindent 
In the new coordinates
$(x,\, p_{x},\, y,\, p_{y})$, this fixed point is described by 
\begin{equation}
\bar{{\cal P}_{2}}: (x=0, p_x=0, y=0, p_y=0).
\end{equation}

\noindent Substituting (\ref{canonica1})--(\ref{canonica4}) into the
hamiltonian (\ref{hamiltoniana}) and linearizing it in the
neighborhood of $\bar{{\cal P}_{2}}$, it can be finally cast in its normal
form
\begin{equation}
H = E_{hyp}+E_{rot}+\epsilon,
\label{normal}
\end{equation}

\noindent where 
\begin{equation}
	\epsilon = mgL
\end{equation}

\noindent fixates the energy surface upon which the motion of the
robot arm will take place. The  energies
\begin{equation}
\left\{
\begin{array}{lllll}
E_{hyp} =& \frac{\displaystyle {\alpha}_{1}}{\displaystyle
	2mL^2}p_{x}^2-\frac{\displaystyle mgL}{\displaystyle
	2}{\alpha}_{3}x^2,\\
& \\
E_{rot} =& \frac{\displaystyle {\alpha}_{2}}{\displaystyle
	2mL^2}p_{x}^2+\frac{\displaystyle mgL}{\displaystyle
	2}{\alpha}_{4}\left(y-\pi\right)^2,
	\\
\end{array}
\right.
\label{energias}\\
\end{equation}

\noindent correspond to the hyperbolic and rotational energies,
respectively. The ${\alpha}_l^{\prime}$s are numerical positive
coefficients originated from the canonical tranformation
(\ref{canonica1}). Those coefficients can be found in appendix B. In
the new coordinates, according to the hamiltonian (\ref{normal}), the
linearized solutions around $\bar{{\cal P}_{2}}$ are
%
\begin{equation}
\left\{
\begin{array}{llllll}
x(t) \cong & \frac{
	\sqrt{2\left(3+\sqrt{17}\right)gL}\left(5+\sqrt{17}\right)}
	{68mgL^2}\left( c_{1}\exp({
	wt})-c_{2}\exp({ -wt})\right)\\
& \\
p_{x}(t)\cong &{ c_{1}\exp({wt})-c_{2}\exp({-wt})}\\
& \\
y(t) \cong & \frac{
	D_{2}\sqrt{2\left(\sqrt{17}-
	3\right)gL}\left(1+5\sqrt{17}\right)}{
	34mgL^2}\mbox{sen}\left(w t + \sigma\right)\\
& \\
p_{y}(t)\cong & D_{2}\cos\left(w t+\sigma\right)\\
\end{array}
\right.
\label{solucoes1}
\end{equation}

\noindent
Notice that in this linear neighborhood the rotational and hyperbolic
motions are totally uncoupled.In (\ref{solucoes1}), $c_{1},\, c_{2},\,
D_{2}$ and $\sigma$ are arbitray constants of integration, depending
on the initial conditions. The frequency $w$ is related to the
frequency $w_{0}$ by 
\begin{equation}
w = \frac{\displaystyle \sqrt{2\left(\sqrt{17}-3\right)}}{\displaystyle 2}w_{0}.
\label{frequenciaw}
\end{equation}

\noindent Similar results are found for the remaining saddle-center
fixed point ${\cal P}_3$, and we shall omit such discussion here. As
to the pure center fixed point, it will be described in the new
coordinates as $\bar{{\cal P}_{1}}:\, 
x=-\frac{\displaystyle 2\pi (5+\sqrt{17})}{\displaystyle
17+5\sqrt{17}},\, 
p_{x}=0,\,
y=\frac{\displaystyle 2\pi
(5+\sqrt{17})}{\displaystyle 17+5\sqrt{17}},\,
p_{y}=0$. 

\vspace{0.3cm}

The linearized solutions in the neighborhood of the pure center are
\begin{equation}
\left\{
\begin{array}{llllll}
x(t) \cong & \frac{\displaystyle -2D_{1}\sqrt{2\left(5+\sqrt{17}\right)gL}\left(17+5\sqrt{17}\right)}{\displaystyle 17mgL^2}\mbox{sen}\left(\Omega_1 t+\theta\right)+\\
& \\
+&D_{2}\sqrt{2\left(5-\sqrt{17}\right)gL}\left(5\sqrt{17}-17\right)\mbox{sen}\left(\Omega_2 t +\sigma\right)\\
& \\
p_{x}(t)\cong & -8D_{1}\left(4+\sqrt{17}\right)\cos\left(\Omega_1 t+\theta\right)+\frac{D_2\left(4+\sqrt{17}\right)}{2}\cos\left(\Omega_2 t+\sigma\right)\\
& \\
y(t) \cong & D_{1}\sqrt{2\left(5+\sqrt{17}\right)gL}\left(153+37\sqrt{17}\right)\mbox{sen}\left(\Omega_1 t+\theta\right)+\\
& \\
+& D_{2}\frac{\sqrt{2\left(5-\sqrt{17}\right)gL}}{34mgL^2}\left(153+37\sqrt{17}\right)\mbox{sen}\left(\Omega_2 t+\sigma\right)\\
& \\
p_{y}(t)\cong & 2D_{1}\cos(\Omega_1 t +\theta)+2D_{2}\cos(\Omega_2 t +\sigma)\\
\end{array}
\right.
\end{equation}

\noindent
where the frequencies 
\begin{displaymath}
\Omega_1=\frac{\sqrt{2(5+\sqrt{17})}}{2} w_0;\, 
\Omega_2=\frac{\sqrt{2(5-\sqrt{17})}}{2} w_0,
\end{displaymath}

\noindent
are also related to the frequencies $w_0$. 

As to the dynamics in the neighborhood of the pure-saddle fixed point
${\cal P}_4$, the solutions are linear combinations of real
exponential functions. We shall restrict our discussion to the
saddle-center fixed points, due to it rich topology.

\subsection{Topology of the linear neighborhood of saddle-center}


\indent Let us now analyze all possible motions in the linear
neighborhood of the saddle-center point. As we have seen, dynamics in
this neighborhood is governed by the hamiltonian (\ref{normal}). Thus,
there are three possibilities: (a) $E_{rot}\neq 0;\, E_{hyp}=0$; (b)
$E_{rot}= 0;\, E_{hyp}\neq 0$ and (c) $E_{rot}\neq 0;\, E_{hyp}\neq
0$. We will discuss each case separately.

\begin{figure}[htb]
\includegraphics[width=16.9 cm,height=9.6 cm]{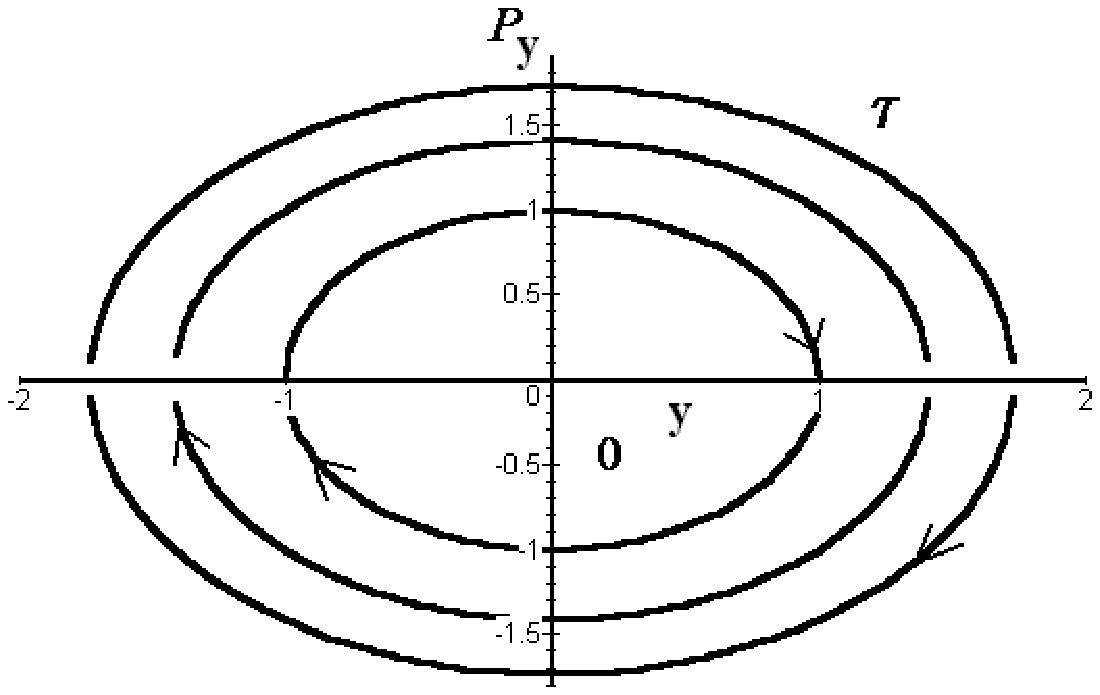}
\caption{\label{f4}\small Projection of orbits upon the
	$(y,P_{y})$ plane near $y = 0$, corresponding to
	unstable periodic orbits in the neighborhood of the fixed point.}
%
\end{figure}

\begin{figure}[htb]
 \includegraphics[width=16.9 cm,height=8.9 cm]{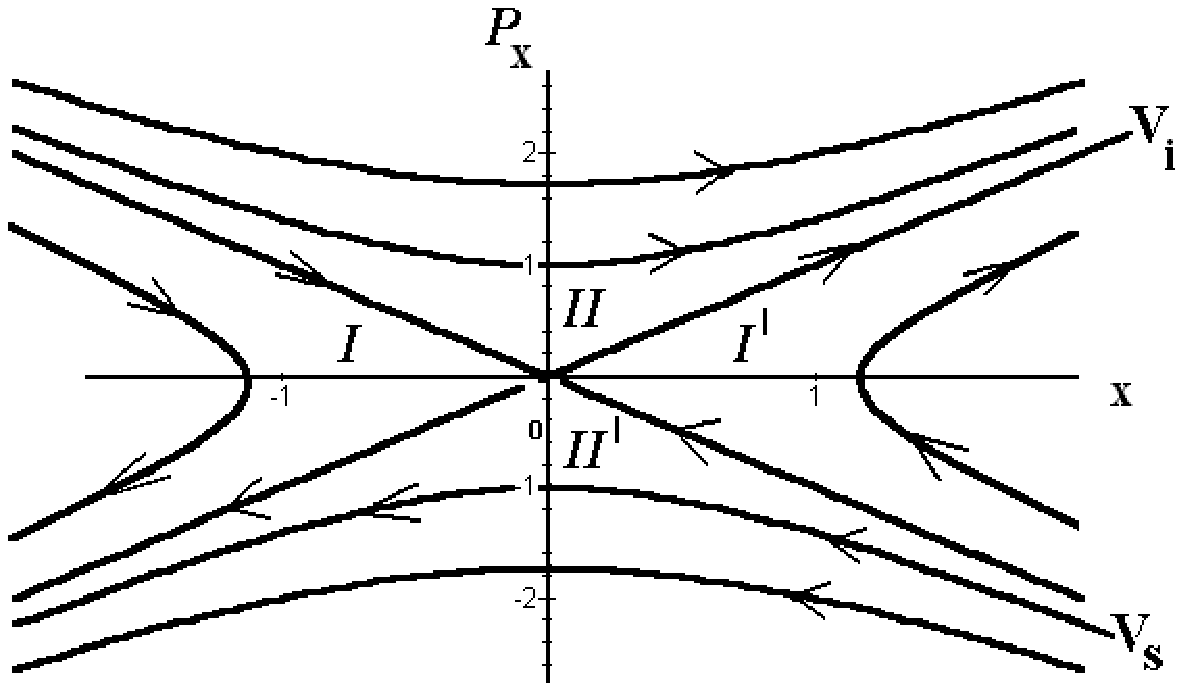}
\caption{\label{f5}\small Projection of orbits upon the $(x,P_{x})$
	plane near $x=0$, revealing the hyperbolic structure in the
	neighborhood of the fixed point.}
%
\end{figure}

\noindent
(a) For $E_{rot}\neq 0,\, E_{hip}= 0$, there are two possibilities:

\begin{enumerate}
\item{ If $(p_{x}=0,\, x=0)$, we have unstable periodic orbits
 $\tau$ upon the $(y,\, p_{y})$ plane, and their projection 
upon the $(x,\, p_{x})$ plane is the 
$(x=0,\, p_{x}=0)$ point. Such orbits depend
continuously on the parameter $\epsilon$, so that

\begin{equation}
E_{rot} = -mgL.
\end{equation}}

\item If $p_{x}=\pm mL\sqrt{\frac{\displaystyle
{\alpha}_{3}g}{\displaystyle {\alpha}_{1}}}x$, the onedimensional
linear manifolds $V_{S}$(stable) and $V_{i}$(unstable) (cf.
Fig.\ref{f4} and Fig.\ref{f5}) tangent the saddle-center are defined.
The separatrices $S$ are non-linear extensions of $V_{S}$ and $V_{i}$.
The general motion is the direct product of periodic orbits $\tau$
with the manifolds $V_{S}$ and $V_{i}$, generating the structures of
cylinders $(\tau \times V_{S})$(stable) and $(\tau \times
V_{i})$(unstable). Orbits on such cylinders have the periodic orbits
$\tau$ as their assymptotic limit $(t \rightarrow \infty)$. Notice
that those cylinders have the same energy as that of the unstable
periodic orbits $\tau$.

\end{enumerate}

\vspace{0.5cm}

\noindent
(b) If $E_{rot}=0$ and $E_{hip}\neq 0$, the motion is upon
the invariant manifold ${\cal M}$ defined by (\ref{manifold1}) or
(\ref{manifold2}).
 
\noindent
The resulting motion of the system consists in hyperbolic orbits on
the $(x,\, p_{x})$ plane, as the projection of orbits on the $(y,\,
p_{y})$ plane is reduced to a single point defined by (\ref{xxc}).
\vspace{0.5cm}

\noindent 
(c) If $E_{rot}\neq 0$ and $E_{hip}\neq 0$, the resultant motion 
is the direct product of hyperbolae related to the
I, I$^{\prime}$, II and II$^{\prime}$ regions of Fig.\ref{f5}, with 
the periodic orbits on the $(y,\, p_{y})$ plane in the
linear neigborhood of the saddle-center fixed point.

\section{The robot arm with constant external torques}

We will analyze now the case of non-null constant external torques.
There are still four fixed points, namely, 
\begin{equation}
{\cal P}^\prime_{1} = 
	\left(
	\begin{tabular}{c}
	$\arctan\left(
		\frac{\beta_1}{\sqrt{-\beta_1^2 + 4 m^2 g^2 L^2}}
		\right)$ \\
	0 \\
	$\arctan\left(
		\frac{\beta_2}{\sqrt{-\beta_2^2 + m^2  g^2 L^2}}
		\right)$ \\
	0 
	\end{tabular}
	\right) ,\ \
{\cal P}^\prime_{2} = 
	\left(
	\begin{tabular}{c}
	$\arctan\left(
		\frac{\beta_1}{\sqrt{-\beta_1^2 + 4 m^2 g^2 L^2}}
		\right)$ \\
	0 \\
	$\arctan\left(
		\frac{\beta_2}{-\sqrt{-\beta_2^2 + m^2  g^2 L^2}}
		\right)$ \\
	0
	\end{tabular}\right), 
\end{equation}	

\begin{equation}
{\cal P}^\prime_{3} = 
	\left(
	\begin{tabular}{c}
	$\arctan\left(
		\frac{\beta_1}{-\sqrt{-\beta_1^2 + 4 m^2 g^2 L^2}}
		\right)$ \\
	0 \\
	$\arctan\left(
		\frac{\beta_2}{\sqrt{-\beta_2^2 + m^2 g^2 L^2}}
		\right)$ \\
	0 
	\end{tabular}\right) ,\ \ 
{\cal P}^\prime_{4} = 
	\left(
	\begin{tabular}{c}
	$\arctan\left(
		\frac{\beta_1}{-\sqrt{-\beta_1^2 + 4 m^2 g^2 L^2}}
		\right)$ \\
	0 \\
	$\arctan\left(
		\frac{\beta_2}{-sqrt{-\beta_2^2+ m^2 g^2 L^2}}
		\right)$ \\
	0 
	\end{tabular}
	\right).
\end{equation}

\noindent where $\beta_1$ and $\beta_2$ stand for the constant
external torques. Coordinates, from top to bottom, are ordered
according to $(\theta_1, p_1, \theta_2, p_2)$. The coordinates of
those fixed points must be real; hence, their existence is subject to
the two simultaneous conditions:
\begin{eqnarray}
|\beta_1| &\leq& 2 m g L, \\
|\beta_2| &\leq&  m g L.
\end{eqnarray}

\noindent If those conditions are satisfied, there will be 
four fixed points, as in the null-torque case. The same procedure
used previously to analyze the nature of such fixed points can be
applied, and their nature determined. It turns out that for
\begin{eqnarray}
|\beta_1| &<& 2 m g L \\
|\beta_2| &<& m g L,
\end{eqnarray}

\noindent 

The nature of fixed points is the same for that in which the torques are null. In
other words, ${\cal P}^\prime_1$ is a pure center, with corresponding energy
\begin{eqnarray}
E_1 &=& 
	-\sqrt{\left(-\beta_2^2 + m^2 g^2 L^2 \right)}
	- \sqrt{-\beta_1^2+4 m^2 g^2 L^2} + \nonumber \\
& & 
	-\arctan\left(\frac{\beta_1}{\sqrt{-\beta_1^2+4 m^2 g^2 L^2}} \right) \beta_1
	-\arctan\left(\frac{\beta_2}{\sqrt{-\beta_2^2+m^2 g^2 L^2}} \right) \beta_2
\end{eqnarray}

\noindent In their turn, ${\cal P}^\prime_2$ and 
${\cal P}^\prime_3$ are saddle-center points, with corresponding energy
\begin{eqnarray}
E_2 &=& 
	-\sqrt{-\beta_2^2 +m^2 g^2 L^2} 
	+\sqrt{-\beta_1^2 +4 m^2 g^2 L^2} + \nonumber \\
& & 
	-\arctan\left(\frac{\beta_1}{-\sqrt{-\beta_1^2+4 m^2 g^2 L^2}}\right) \beta_1
	-\arctan\left(\frac{\beta_2}{\sqrt{-\beta_2^2 + m^2 g^2 L^2}}\right)  \beta_2
\end{eqnarray}

\noindent and

\begin{eqnarray}
E_3 &=& 
	\sqrt{-\beta_2^2 + m^2 g^2 L^2}
	-\sqrt{-\beta_1^2+ 4 m^2 g^2 L^2} +
	\nonumber \\
& & 
	-\arctan\left(\frac{\beta_1}{\sqrt{-\beta_1^2 + 4 m^2 g^2 L^2}}\right) \beta_1
	-\arctan\left(\frac{\beta_2}{-\sqrt{-\beta_2^2+ m^2 g^2 L^2}}\right) \beta_2
\end{eqnarray}

\noindent respectively. The fourth fixed point ${\cal P}^\prime_4$ is
a pure saddle, with energy

\begin{eqnarray}
E_4 &=& 
	\sqrt{-\beta_2^2+m^2 g^2 L^2}
	+\sqrt{-\beta_1^2+4 m^2 g^2 L^2}
	+ \nonumber \\
& & 
	-\arctan\left(\frac{\beta_1}{-\sqrt{-\beta_1^2+4 m^2 g^2 L^2}}\right) \beta_1
	-\arctan\left(\frac{\beta_2}{-\sqrt{-\beta_2^2+m^2 g^2 L^2}}\right) \beta_2.
\end{eqnarray}

\noindent If $\beta_1=\beta_2=0$, we fall into the null torques case,
and the classification of those points is the same. In the special
case
\begin{equation}
|\beta_2| = m g L  \mbox{ and } |\beta_1| = 2 m g L
\end{equation}

\noindent it can be shown that all fixed points are degenerate, i.e.,
the jacobian matrices of the linearized system in the neighborhood of
those fixed points have all vanishing eigenvalues. In this case, a new
procedure must be taken, so that at first this degenerescence is
raised, and then the points are classified\cite{bogo}. We will not
approach the degenerate case here.

From what has been seen above, the topology of orbits in the {\it
linear} neighborhood of fixed points is the same for both the
null-torques and non-null constant torques cases.

By the introduction of constant external torques, their intensity can
be adjusted so that the robot arm be held any desired configuration in
equilibrium. The majority of those equilibrium points are
unstable, though: any fluctuation on the initial conditions
can take the system out of equilibrium. In the case of the
saddle-center points, the choice of certain branches of the saddle can
lead to stable equilibrium, even in the presence of fluctuations.


\section{Conclusions and final remarks}

We have proposed that the planar manipulator with two links and two
rotational joins, a system with two degrees of freedom, be described
by the hamiltonian of a double pendulum subject to two external
torques. In its phase space, four fixed points (stationary solutions)
can be found; a pure center, a pure saddle, and two saddle-centers,
which can be used as a clue to the structure of orbits in in all phase
space. We have also observed that there are two similar invariant
manifolds, each one of them associated to a saddle-center points. The
phase portrait of the system upon those manifolds resembles that of a
mathematical pendulum of arbitrary amplitude. Dynamics upon such
manifolds is governed by a unidimensional autonomous system, thus
being totally integrable. For any set of initial conditions placed on
one such manifold, the orbits will be confined to that manifold. We
should keep in mind, however, that those manifolds are embedded into a
four-dimensional phase space, so that the system in general in
non-integrable.

An important result is that for constant external torques, the phase
space topology in the linear neighborhood of the fixed points is the
same as that of the null-torques case, if certain conditions involving
the torques and parameters of the system are satisfied. With a proper
choice of torque intensities, then, we may define any point in
configuration space as a fixed point, without altering its nature, in
relation to the null-torques case. In particular, the topology of
orbits in the linear neighborhood of the saddle-center points is the
same, as the existence of the corresponding invariant manifolds.

A possible continuation of this work involves the analysis of the
non-linear neighborhood of fixed points. Strong indications of chaotic
behavior are expected, due to the non-integrability of the system.

\section*{Acknowledgment}

G. A Monerat \& E. V. Corr\^{e}a Silva thank the Conselho Nacional de Desenvolvimento Cient\'{\i}fico e Tecnol\'ogico (CNPq), Brazil, for finantial support.



\end{document}